\documentclass{article}
\usepackage[utf8]{inputenc}
\usepackage{graphicx}
\usepackage{caption}
\usepackage{subcaption}
\usepackage{geometry}
\usepackage{enumitem}
\usepackage{booktabs}
\usepackage{array}
\usepackage{placeins}
\usepackage{float}
\newcolumntype{P}[1]{>{\centering\arraybackslash}p{#1}}
\title{\textbf{ALOHA2 Robot Kitchen Application Scenario Reproduction Report}}
\author{
    \textbf{Haoyang Wu}$^1$, \textbf{Siheng Wu}$^1$ \\
    \textbf{Dr. William X. Liu}$^2$, \textbf{Fangui Zeng}$^2$ \\
    $^1$Peking University \\
    $^2$Towngaslifestyle Information Service Co., Ltd.
}
\date{}

\begin{document}

\maketitle

\section*{A. Basic Introduction to the ALOHA2 System}

\subsection*{I. Overview}
\begin{figure}[htbp]
    \centering
    \includegraphics[width=0.6\textwidth]{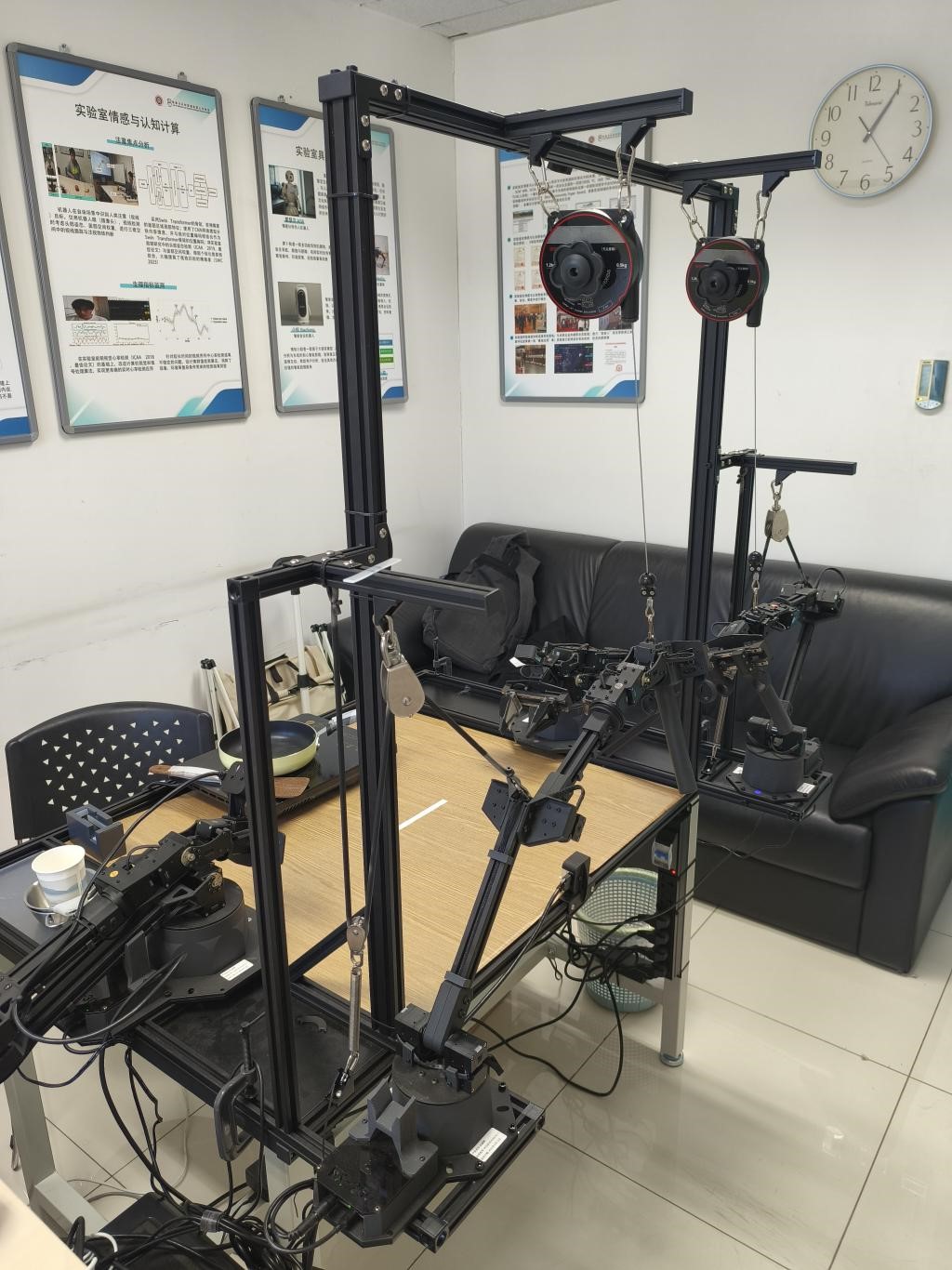}
    \caption{ALOHA2 Overview}
    \label{fig:overview}
\end{figure}

ALOHA2 is an enhanced version of the dual-arm teleoperated robot ALOHA \cite{zhao2024aloha2}, featuring higher performance and robustness compared to the original design \cite{zhao2023aloha}, while also being more ergonomic. Like ALOHA, ALOHA2 consists of two grippers and two ViperX 6-DoF arms, as well as two smaller WidowX arms. Users control the follower mechanical arms by operating the leader mechanical arms through back-driving. The device also includes cameras that generate images from multiple viewpoints, allowing for RGB data collection during teleoperation. The robot is mounted on a 48-inch $\times$ 30-inch table, equipped with an aluminum frame that provides additional mounting points for cameras and gravity compensation systems (as shown in Figure \ref{fig:overview}).

\subsection*{II. Improvements over ALOHA}
\textbf{1. Improvement Overview}
\begin{itemize}[noitemsep,topsep=0pt]
    \item \textbf{Work Performance and Task Range:} ALOHA2 seeks to enhance key components, including grippers and controllers, to improve ALOHA2's performance over ALOHA for achieving a broader set of operational tasks.
    \item \textbf{User-Friendly and Ergonomic:} To optimize data collection at scale, ALOHA2 prioritizes user experience and comfort. This includes improvements in responsiveness and ergonomics for user-facing system design.
    \item \textbf{Robustness:} ALOHA2 aims to improve system robustness to minimize downtime and repairs caused by diagnostics. This includes simplifying mechanical design and ensuring overall ease of use.
\end{itemize}

\textbf{2. Specific Improvements}
\begin{enumerate}[label=\alph*.]
    \item \textbf{Active Gripper:} For smoother teleoperation and improved ergonomics, ALOHA2 replaced the original ALOHA design's scissor-head gripper with a low-friction rail design, reducing mechanical complexity (see Figure \ref{fig:gripper_comp}). To further reduce strain, ALOHA2 also reduced back-driving friction by replacing the original active gripper motor (XL430-W250-T) with a low-friction alternative (XC430-W150-T) that has a lower gear ratio. The new design uses low-friction metal gears instead of plastic gears, with opening and closing forces 10 times smaller than the previous ALOHA scissor design. This significantly reduces operator hand fatigue and strain during long data collection sessions, especially for the lumbrical muscles responsible for opening the gripper.
    
    \begin{figure}[htbp]
        \centering
        \includegraphics[width=0.6\textwidth]{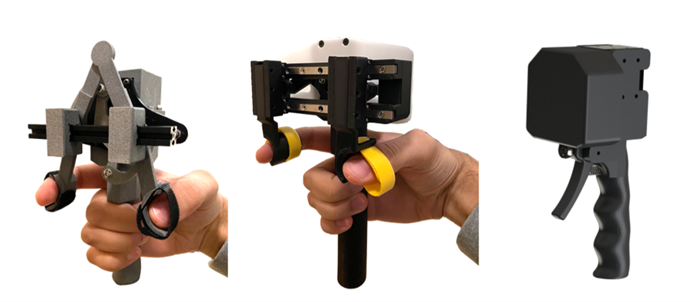}
        \caption{Comparison of Active Grippers between ALOHA2 and ALOHA1}
        \label{fig:gripper_comp}
    \end{figure}
    
    \item \textbf{Follower Gripper:} ALOHA2 designed and manufactured low-friction follower claws to replace the original ALOHA design. The lower friction reduces perceived delay between leader and follower grippers, significantly improving user experience in teleoperation (see Figure \ref{fig:follower_gripper}). ALOHA2 demonstrates the delay difference between old and new designs in supplementary videos. The new gripper can apply more than twice the force of the old design, allowing for stronger and more stable object grasping.
    
    Additionally, ALOHA2 improved the compliance of the gripping mechanism by removing the original PLA + acrylic structure and replacing it with 3D-printed carbon fiber nylon. Both the gripping fingers and support structure can deform under force, improving system safety.
    
    ALOHA2 retained the original ALOHA's "transparent" finger connection design. Furthermore, ALOHA2 improved the gripping tape on the fingers. ALOHA2 found that the original gripping tape on ALOHA wore out over time, and the rounded fingertips made it difficult to pick up small objects. ALOHA2 uses polyurethane gripping tape on the interior of the gripper. ALOHA2 also applies tape to the exterior of the fingers to increase traction for tasks requiring manipulation of objects with the gripper's exterior.
    
    \begin{figure}[htbp]
        \centering
        \includegraphics[width=0.5\textwidth]{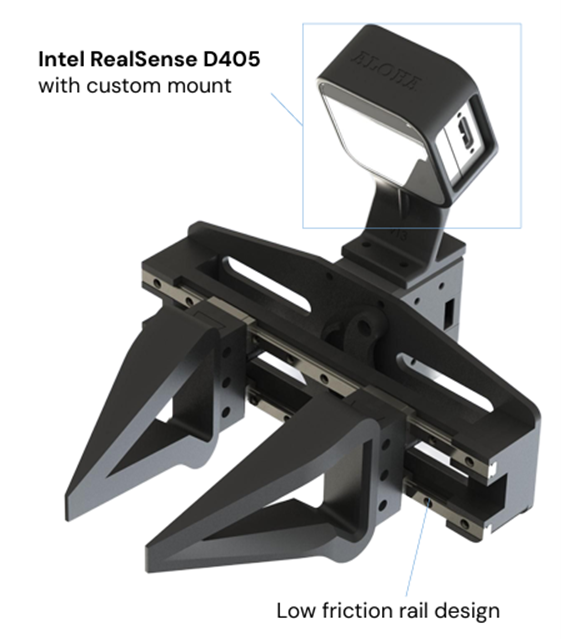}
        \caption{Follower Gripper and Its Mounted Camera}
        \label{fig:follower_gripper}
    \end{figure}
    
    \item \textbf{Gravity Compensation:} ALOHA2 designed a more robust passive gravity compensation for the active mechanical arms to reduce operator burden during teleoperation. ALOHA2 uses off-the-shelf components, including adjustable suspended retractors, allowing operators to adjust load balancing force to their comfort level.
    
    \item \textbf{Support Frame:} ALOHA2 designed a support frame and built it using 20x20mm aluminum profiles (see Figure \ref{fig:support_frame}). The frame provides support for active mechanical arms and gravity compensation systems, and provides mounting points for overhead and bird's-eye cameras. Compared to ALOHA, ALOHA2 simplified the design by removing the vertical frame on the side of the table opposite the teleoperator. The added space allows for different styles of data collection. For example, human collaborators can more easily stand on the other side of the workspace and interact with the robot, allowing for human-robot interaction data collection. Additionally, larger props can be placed in front of the table for the robot to interact with.
    
    \begin{figure}[htbp]
        \centering
        \includegraphics[width=0.4\textwidth]{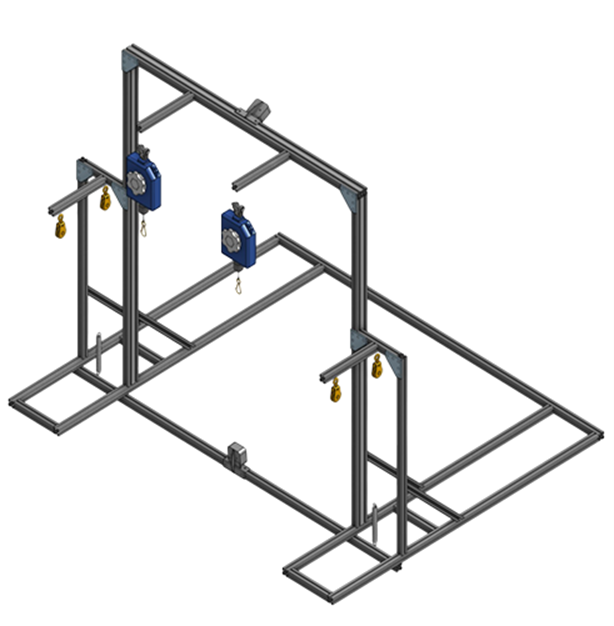}
        \caption{Support Frame Diagram}
        \label{fig:support_frame}
    \end{figure}
    
    \item \textbf{Cameras:} ALOHA2 upgraded the cameras used in the ALOHA system to 4 RealSense D405 cameras (see Figure \ref{fig:follower_gripper}). These cameras achieve high-resolution RGB and depth in a very small size, as well as providing global shutters. Note that while depth and global shutters are not necessary for the results demonstrated on the first ALOHA system, they might be considered "nice to have" for enabling different experiments and pushing performance. ALOHA2 designed new camera mounts for wrist cameras, overhead cameras, and bird's-eye cameras. The lower camera on the wrist reduces the number of collision states and improves teleoperation for certain fine-grained manipulation tasks, particularly those requiring close contact between arms or navigation in tight spaces.
\end{enumerate}

\subsection*{III. ALOHA2 Component Description}
\begin{itemize}
    \item \textbf{Active Mechanical Arms:} Operators control the follower mechanical arms by operating the active mechanical arms to complete demonstrations. The active mechanical arms adopt a low-friction track design that can reduce operator hand fatigue (see Figure \ref{fig:active_arms}).
    \begin{figure}[htbp]
        \centering
        \includegraphics[width=0.5\textwidth]{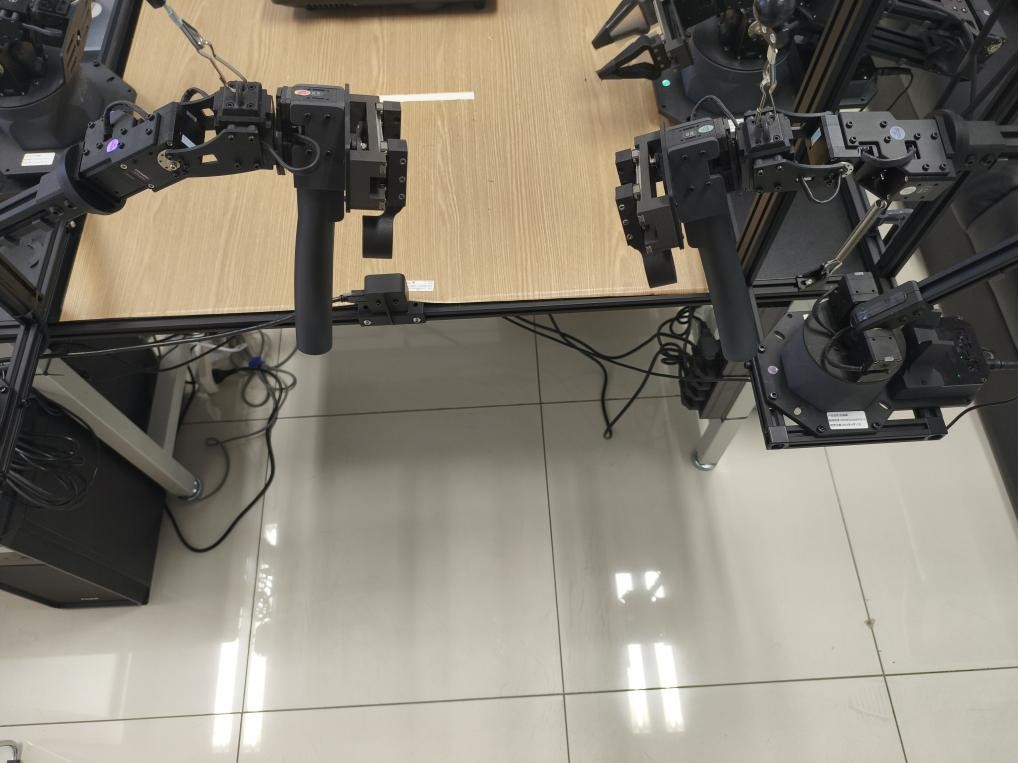}
        \caption{Active Mechanical Arms}
        \label{fig:active_arms}
    \end{figure}
    
    \item \textbf{Follower Mechanical Arms:} The mechanical arms that ALOHA2 uses to actually interact with items and complete tasks, capable of bearing certain loads. Anti-slip tape is wrapped around the gripper to improve friction and better grasp objects (see Figure \ref{fig:follower_arms}).
    \begin{figure}[htbp]
        \centering
        \includegraphics[width=0.5\textwidth]{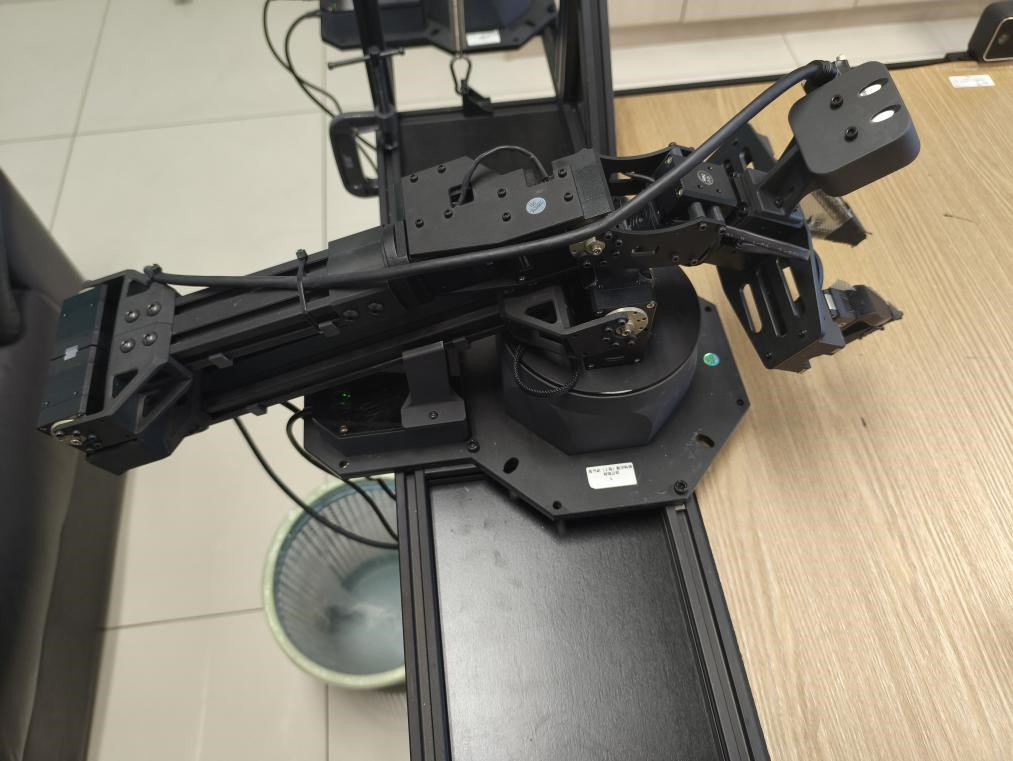}
        \caption{Follower Mechanical Arms (Left Arm)}
        \label{fig:follower_arms}
    \end{figure}
    
    \item \textbf{Gravity Compensation Device:} The gravity compensation device can reduce the operator's burden, including adjustable suspended springs that allow operators to adjust load balancing force according to their comfort level (see Figure \ref{fig:gravity_comp}).
    \begin{figure}[htbp]
        \centering
        \includegraphics[width=0.5\textwidth]{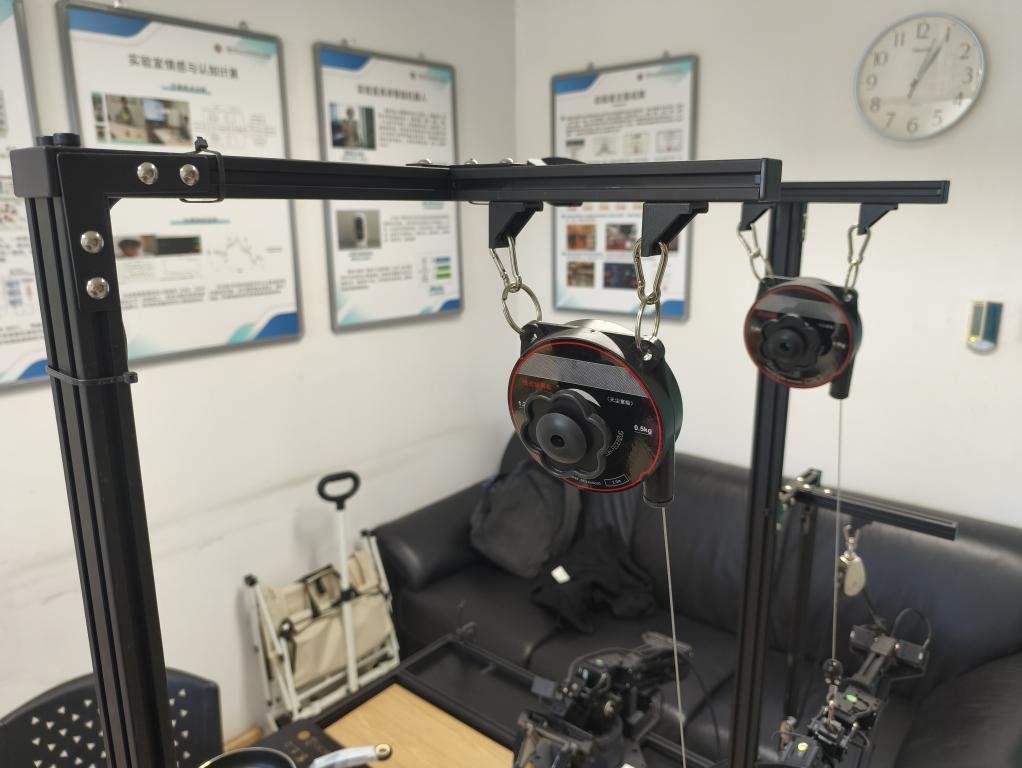}
        \caption{Gravity Compensation Device}
        \label{fig:gravity_comp}
    \end{figure}
    
    \item \textbf{Support Frame:} Made from 20x20mm aluminum materials, providing support for active mechanical arms and gravity compensation devices, and mounting points for overhead cameras.
    
    \item \textbf{Cameras:} Used to record RGB data from teleoperation demonstrations and provide image information during inference. Four Intel RealSense D405 cameras are placed at the left and right follower mechanical arms, top, and bottom respectively (see Figure \ref{fig:camera}).
    \begin{figure}[htbp]
        \centering
        \includegraphics[width=0.5\textwidth]{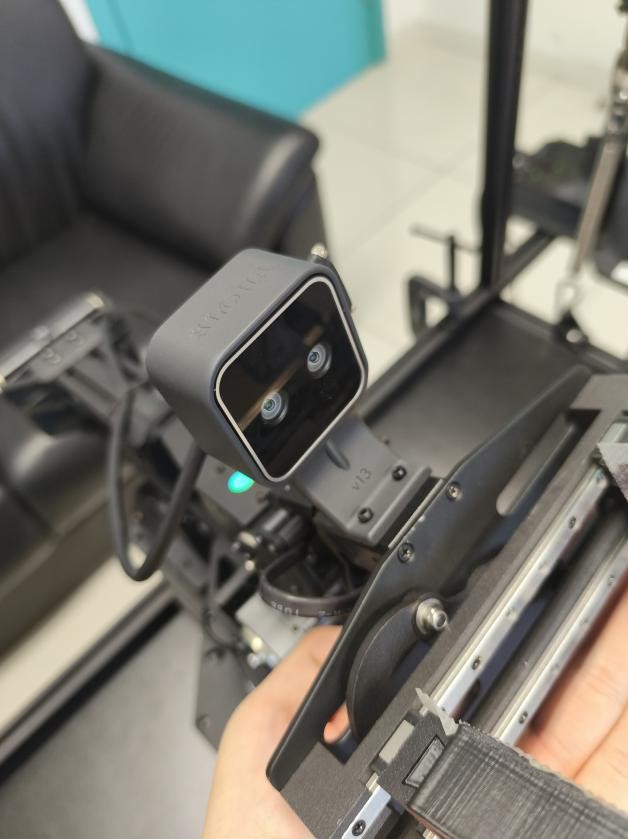}
        \caption{Camera}
        \label{fig:camera}
    \end{figure}
\end{itemize}

\section*{B. Experimental Setup}

\subsection*{I. Task Introduction}
Our goal is to achieve using ALOHA2 mechanical arms to complete a cooking task using an induction cooker. The main task is to put ingredients into a pan, then use the induction cooker to fry the food, and finally pour the food out onto a plate. This work builds upon recent advances in robot learning \cite{brohan2022rt1, brohan2023rt2, openxembodiment2024dataset} and robotic cooking \cite{dann2021robot}.

\subsection*{II. Experimental Props}    
\begin{enumerate}
    \item \textbf{Ingredients} \\
    We used both toy ingredients and real ingredients as experimental props. The toy ingredients are plastic toy ingredients of various colors. To facilitate cleaning, we mainly used toy ingredients when collecting data and testing models. For real ingredients, we chose frozen scallop meat, which we used in final testing to determine how ALOHA2 performs in real scenarios (Figures \ref{fig:toy_ingredients} and \ref{fig:scallop}).
    \begin{figure}[htbp]
        \centering
        \includegraphics[width=0.5\textwidth]{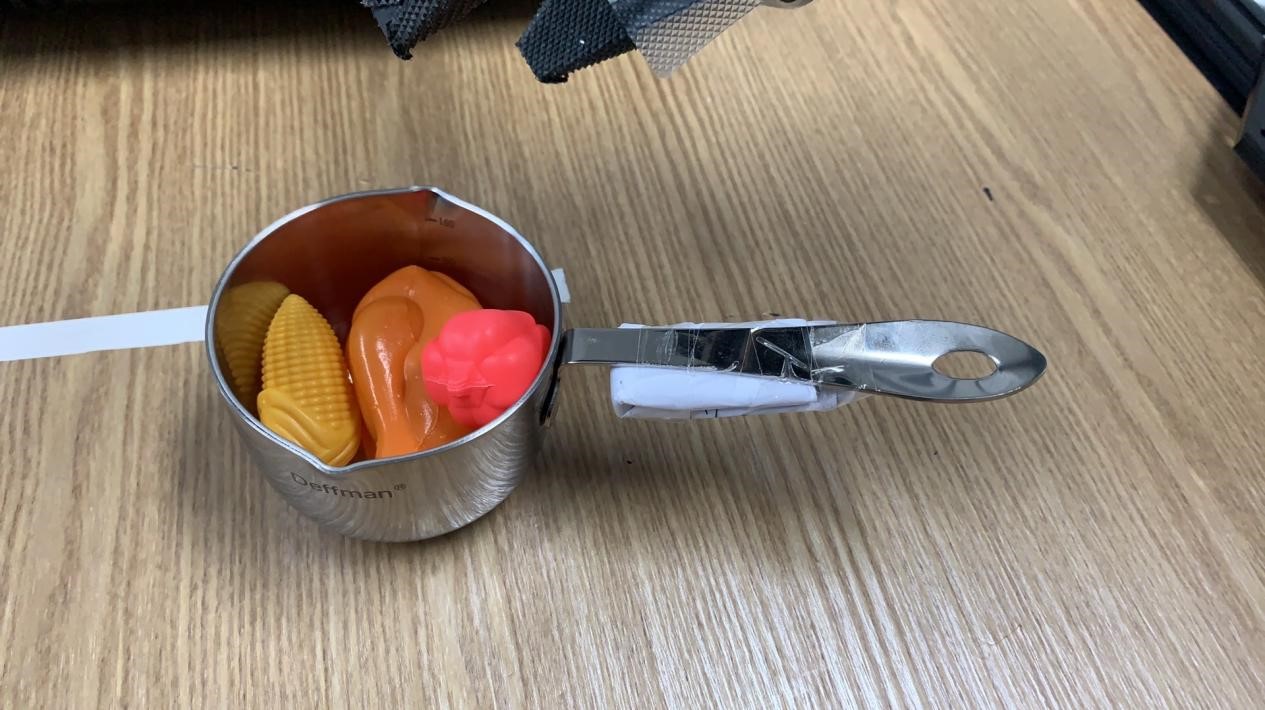}
        \caption{Toy Ingredients}
        \label{fig:toy_ingredients}
    \end{figure}
    \begin{figure}[H]
        \centering
        \includegraphics[width=0.5\linewidth]{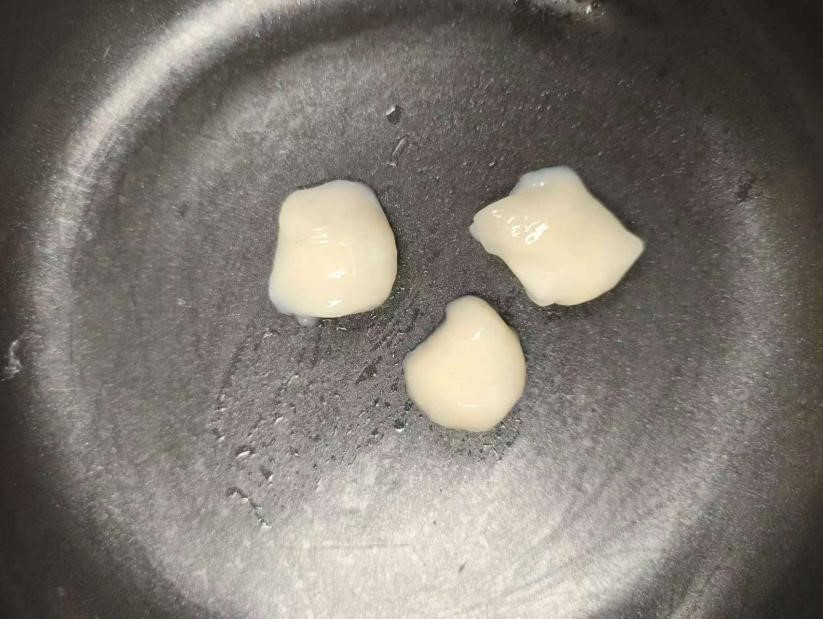}
        \caption{Frozen Scallop Meat}
        \label{fig:scallop}
    \end{figure}

    \item \textbf{Pan} \\
    A metal pan with a black handle and interior, and tea green exterior (Figure \ref{fig:pan}).
    \begin{figure}[H]
        \centering
        \includegraphics[width=0.4\textwidth]{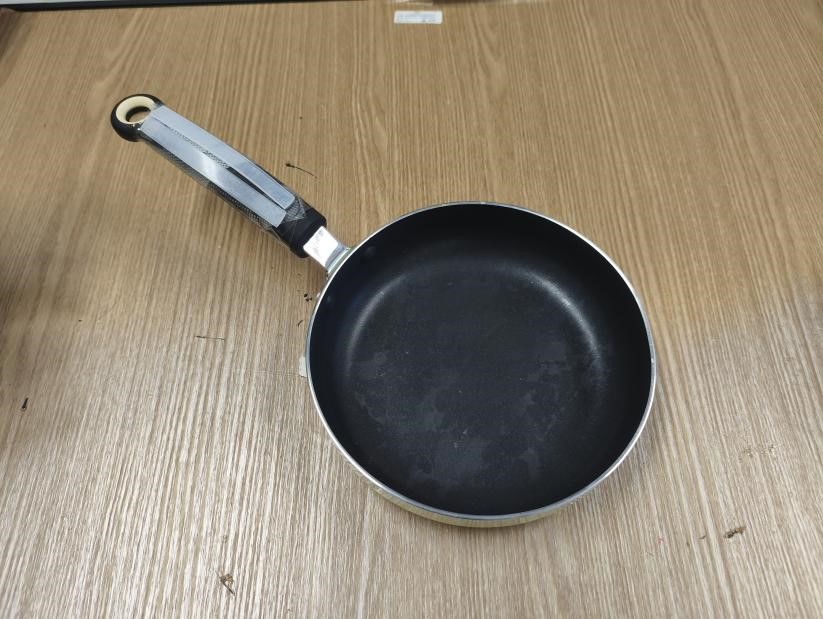}
        \caption{Pan}
        \label{fig:pan}
    \end{figure}
    
    \item \textbf{Spatula} \\
    A wooden spatula (Figure \ref{fig:spatula}).
    \begin{figure}[H]
        \centering
        \includegraphics[width=0.3\textwidth]{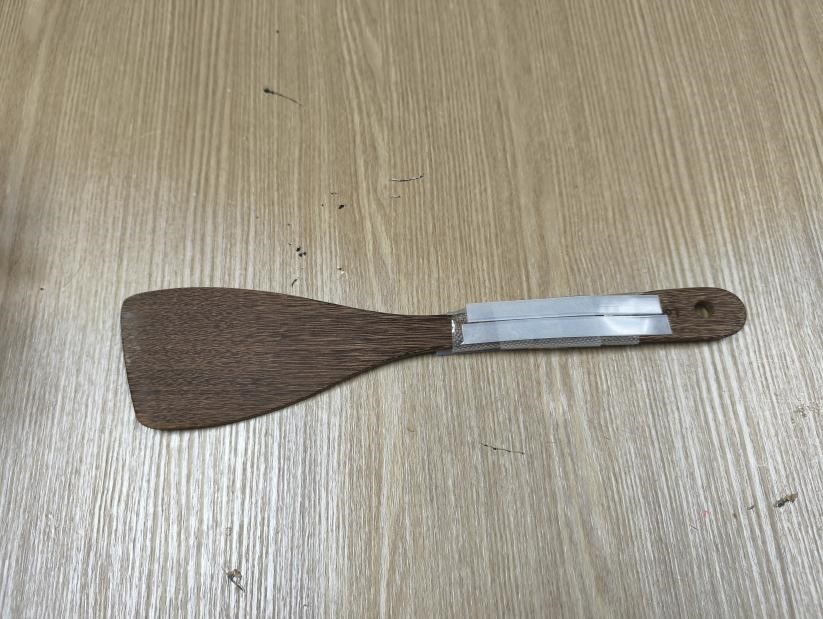}
        \caption{Spatula}
        \label{fig:spatula}
    \end{figure}
    
    \item \textbf{Plate} \\
    A white disposable paper plate (Figure \ref{fig:plate}).
    \begin{figure}[H]
        \centering
        \includegraphics[width=0.3\textwidth]{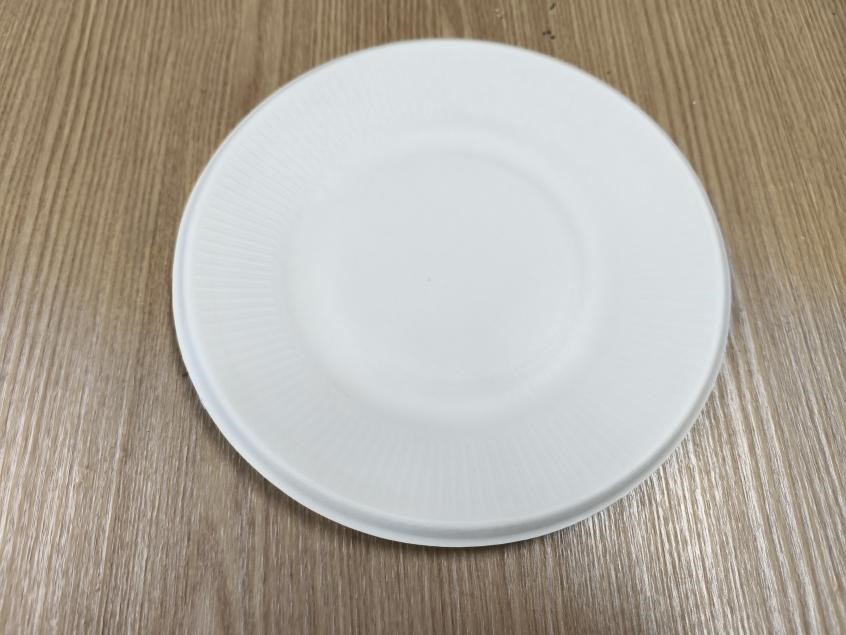}
        \caption{White Paper Plate}
        \label{fig:plate}
    \end{figure}
    
    \item \textbf{Induction Cooker} \\
    An induction cooker used to heat food (Figure \ref{fig:cooker}).
    \begin{figure}[H]
        \centering
        \includegraphics[width=0.4\textwidth]{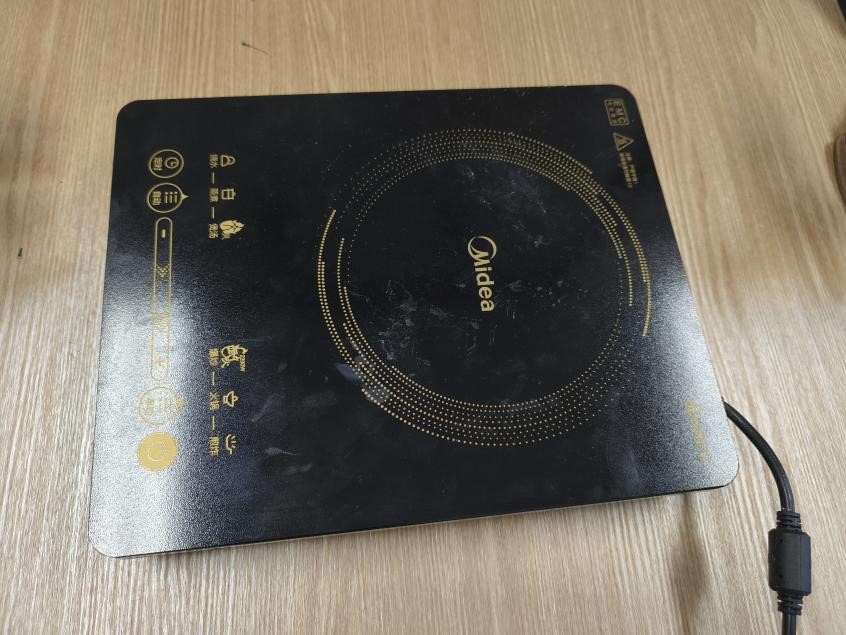}
        \caption{Induction Cooker}
        \label{fig:cooker}
    \end{figure}
    
    \item \textbf{Silver Small Pot} \\
    A metal silver long-handled small pot used to hold initial raw ingredients (Figure \ref{fig:pot}).
    \begin{figure}[H]
        \centering
        \includegraphics[width=0.4\textwidth]{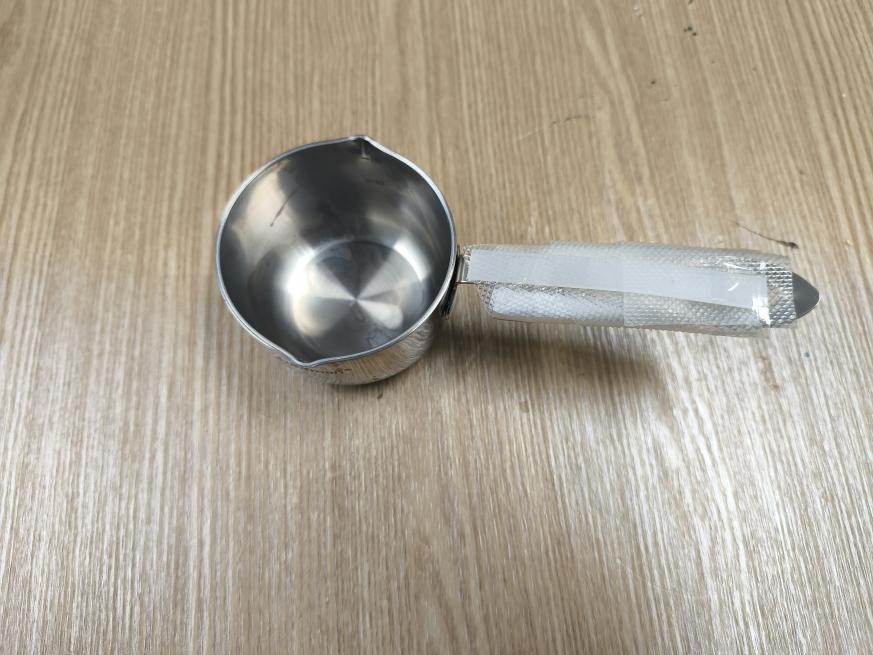}
        \caption{Silver Small Pot}
        \label{fig:pot}
    \end{figure}
\end{enumerate}

\section*{C. Detailed Description of Four Experimental Stages}

\subsection*{I. Pan Movement in Cluttered Scenes}
\textbf{1. Action Flow} \\
The pan is initially located on the left side of the table, with the handle facing right. The right mechanical arm grasps the pan handle, places the pan smoothly on the induction cooker, and releases the handle. Due to the large amount of clutter on the table, how to avoid interference from clutter on positioning and how to avoid these obstacles during pan grasping is the difficulty of this task.

\begin{figure}[htbp]
    \centering
    \includegraphics[width=0.8\linewidth]{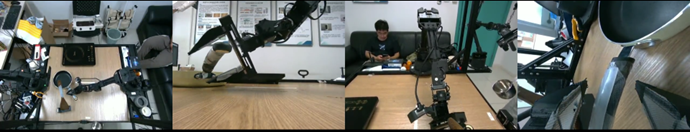}
    \caption{Pan movement action sequence step 1}
    \label{fig:pan_movement_1}
\end{figure}
\begin{figure}[htbp]
    \centering
    \includegraphics[width=0.8\linewidth]{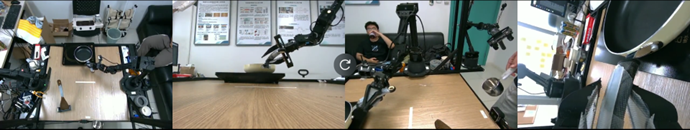}
    \caption{Pan movement action sequence step 2}
    \label{fig:pan_movement_2}
\end{figure}

\textbf{2. Data Collection} \\
A total of 50 samples were collected, each lasting 10 seconds. During collection, attention must be paid to finding the exact position of the pan handle and extending the gripper to the inner part of the handle so that the handle tail is stuck on the gripper, making grasping more stable. The pan is placed on the left side of the table for easy grasping by the right mechanical arm; the induction cooker is placed on the upper part of the table. Since it has little impact on results, no ingredients were added to the pan during collection.

\textbf{3. Model Training} \\
Using 50 data samples for 8000 training iterations.

\textbf{4. Inference Execution} \\
The best-performing model was selected for final inference. 10 trials were conducted with an 80\% success rate. In a few cases, the pan couldn't be grasped or was dropped due to unstable grasping. It's worth noting that although the mechanical arm can successfully grasp the pan and place it above the induction cooker, only once did it retract the arm; in other cases, it maintained the placement posture to end the task.

\begin{itemize}
    \item Success: 
    \begin{figure}[H]
        \centering
        \includegraphics[width=0.8\linewidth]{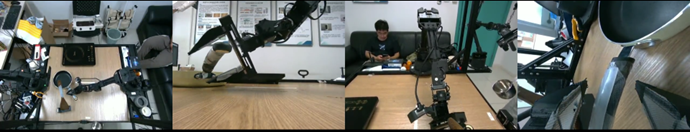}
        \caption{Successful pan grasping}
        \label{fig:pan_success_1}
    \end{figure}
    \begin{figure}[H]
        \centering
        \includegraphics[width=0.8\linewidth]{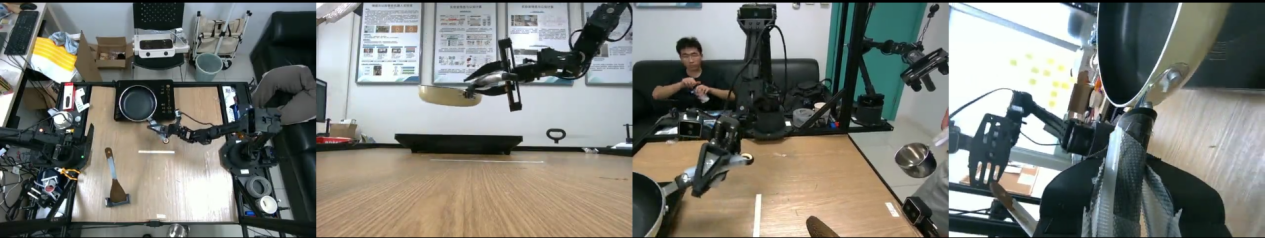}
        \caption{Successful pan placement}
        \label{fig:pan_success_2}
    \end{figure}
    \begin{figure}[H]
        \centering
        \includegraphics[width=0.8\linewidth]{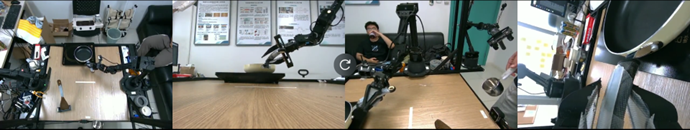}
        \caption{Successful completion}
        \label{fig:pan_success_3}
    \end{figure}
    
    \item Failure: 
    \begin{figure}[H]
        \centering
        \includegraphics[width=0.8\linewidth]{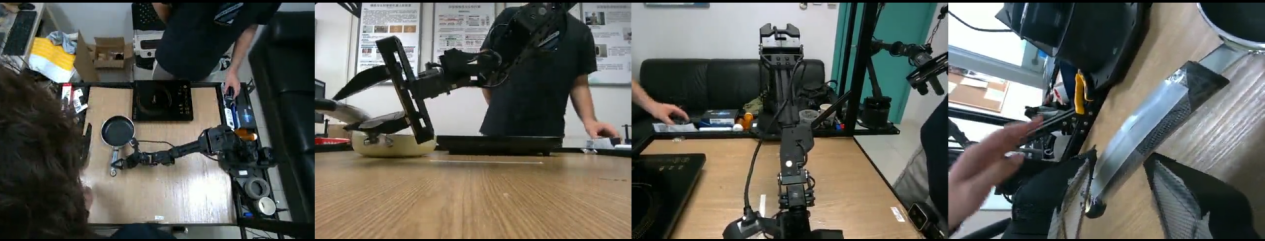}
        \caption{Failed grasping attempt}
        \label{fig:pan_failure_1}
    \end{figure}
    \begin{figure}[H]
        \centering
        \includegraphics[width=0.8\linewidth]{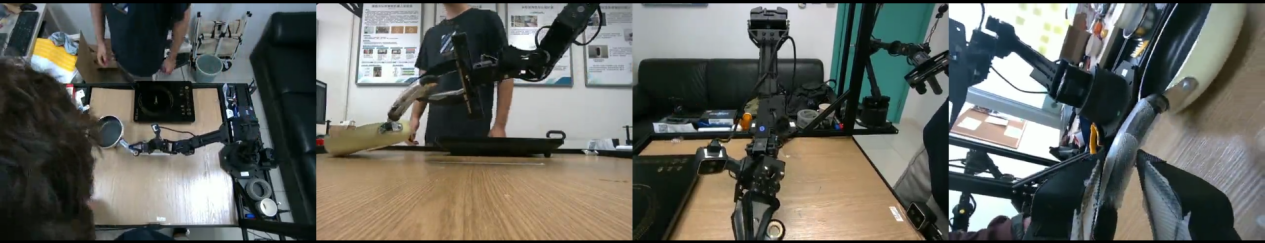}
        \caption{Unstable grasping}
        \label{fig:pan_failure_2}
    \end{figure}
    \begin{figure}[H]
        \centering
        \includegraphics[width=0.8\linewidth]{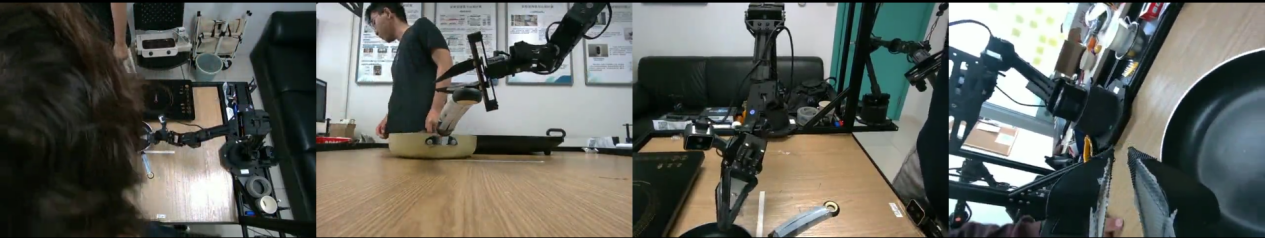}
        \caption{Dropped pan}
        \label{fig:pan_failure_3}
    \end{figure}
\end{itemize}

\textbf{Cause Analysis:} This may be because the pan placement position was relatively fixed during collection, and slight changes during inference could lead to grasping failures; additionally, if the gripper cannot catch the pan handle, unstable grasping situations may occur.

\begin{figure}[H]
    \centering
    \includegraphics[width=0.5\linewidth]{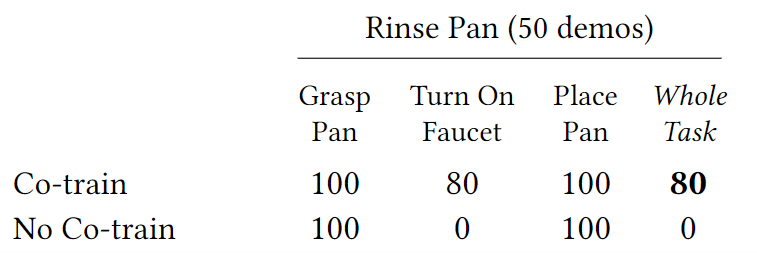}
    \caption{The success rate of original paper's pan grasping and placement}
    \label{fig:paper_success1}
\end{figure}

\begin{figure}[H]
    \centering
    \includegraphics[width=0.5\linewidth]{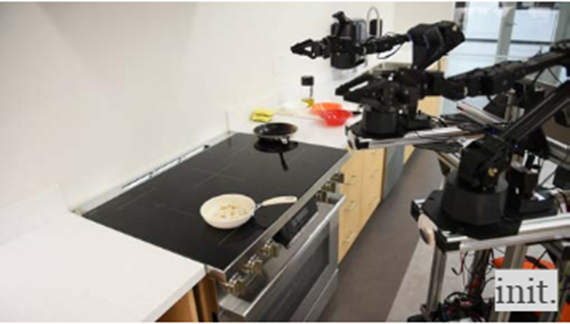}
    \caption{The scene of grasping and placing a frying pan in the original paper}
    \label{fig:paper_scene1}
\end{figure}

\textbf{5. Comparison with Original Paper} 
The original paper's pan grasping and placement task had a 100\% success rate (see Figure \ref{fig:paper_success1}), higher than our current success rate. However, since the original paper did not test in cluttered scenes (see Figure \ref{fig:paper_scene1}), we consider this result reasonable.

\textbf{6. Summary}
\begin{enumerate}[label=\roman*.]
    \item Currently, the completion rate for grasping and placement is relatively high, but the pan needs to be placed in a comfortable position. If placed randomly, the success rate will be reduced to some extent.
    \item The mechanical arm cannot be retracted in time after placing the pan. If the next stage of inference continues, the arm may hit the pan handle when retracting.
    \item Subsequently, while ensuring stability, movement speed and completeness should be improved. For example, during collection, the pan position should be slightly varied, and each grasp should be accurate.
\end{enumerate}

\subsection*{II. Pouring Ingredients from Metal Small Pot into Pan}
\textbf{1. Action Flow} \\
The small pot is initially located in the left area of the table with its handle facing right. The right robotic arm grasps the handle of the small pot, steadily moves it above the frying pan, pours the food into the frying pan, and then places the small pot back on the table smoothly.

\begin{figure}[H]
    \centering
    \includegraphics[width=0.8\linewidth]{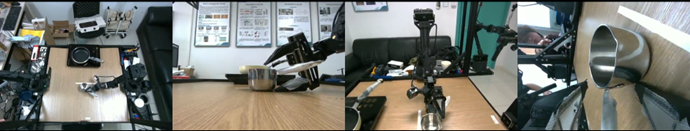}
    \caption{Pouring ingredients step 1}
    \label{fig:pouring_1}
\end{figure}
\begin{figure}[H]
    \centering
    \includegraphics[width=0.8\linewidth]{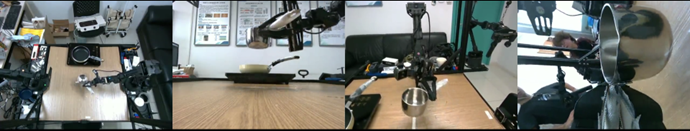}
    \caption{Pouring ingredients step 2}
    \label{fig:pouring_2}
\end{figure}
\begin{figure}[H]
    \centering
    \includegraphics[width=0.8\linewidth]{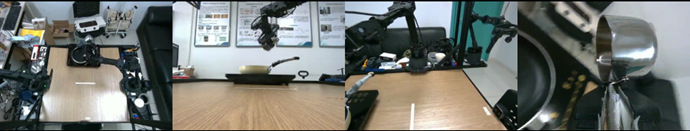}
    \caption{Pouring ingredients step 3}
    \label{fig:pouring_3}
\end{figure}
\begin{figure}[H]
    \centering
    \includegraphics[width=0.8\linewidth]{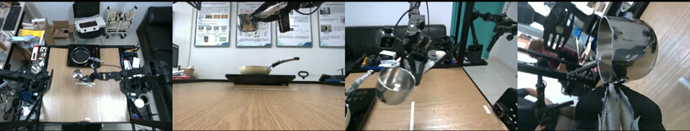}
    \caption{Pouring ingredients step 4}
    \label{fig:pouring_4}
\end{figure}

\textbf{2. Data Collection} \\
A total of 100 data samples were collected in two batches, with improvements made in the latter 50 samples. Each data sample has a duration of 15 seconds. During collection, the small pot was placed on the left side of the table, and the frying pan was placed on the induction cooker. It should be noted that the robotic arm's gripper should grasp the front section of the small pot's handle to allow the tail to be stuck, thereby improving stability. When pouring, the pot must be completely flipped to ensure all contents are poured out.

\textbf{3. Model Training} \\
Two training sessions were conducted successively, using 100 data samples for 8000 training iterations.

\textbf{4. Execution and Inference} \\
The best-performing model was selected for final inference. 10 experiments were conducted, with a success rate of 80\%. There was no problem in locating the small pot or pouring, but there were relatively high proportions of cases where the small pot was dropped into the frying pan after pouring and cases of repeated pouring.

\begin{itemize}
    \item Success: 
    \begin{figure}[H]
        \centering
        \includegraphics[width=0.8\linewidth]{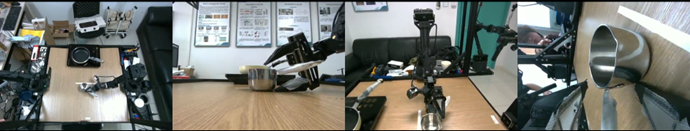}
        \caption{Successful pouring step 1}
        \label{fig:pouring_success_1}
    \end{figure}
    \begin{figure}[H]
        \centering
        \includegraphics[width=0.8\linewidth]{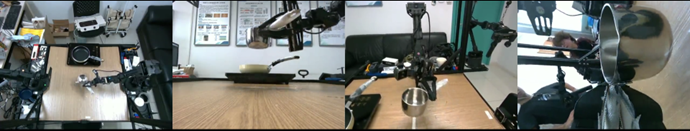}
        \caption{Successful pouring step 2}
        \label{fig:pouring_success_2}
    \end{figure}
    \begin{figure}[H]
        \centering
        \includegraphics[width=0.8\linewidth]{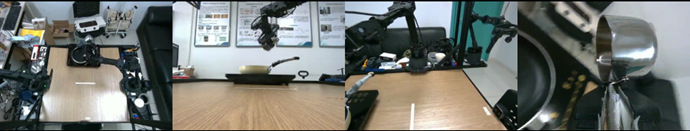}
        \caption{Successful pouring step 3}
        \label{fig:pouring_success_3}
    \end{figure}
    \begin{figure}[H]
        \centering
        \includegraphics[width=0.8\linewidth]{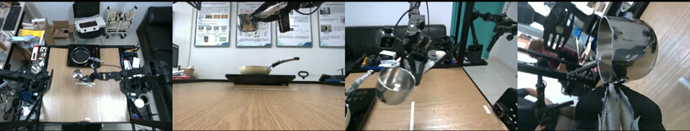}
        \caption{Successful completion}
        \label{fig:pouring_success_4}
    \end{figure}
    
    \item Failure(Continuous pouring): 
    \begin{figure}[H]
        \centering
        \includegraphics[width=0.8\linewidth]{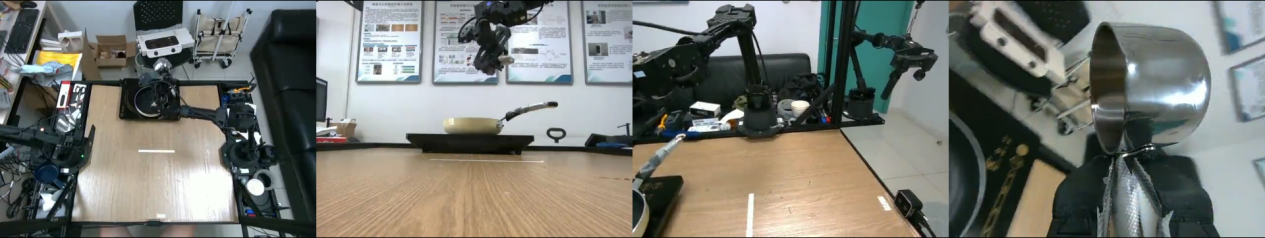}
        \caption{Continuous pouring issue 1}
        \label{fig:pouring_failure_1}
    \end{figure}
    \begin{figure}[H]
        \centering
        \includegraphics[width=0.8\linewidth]{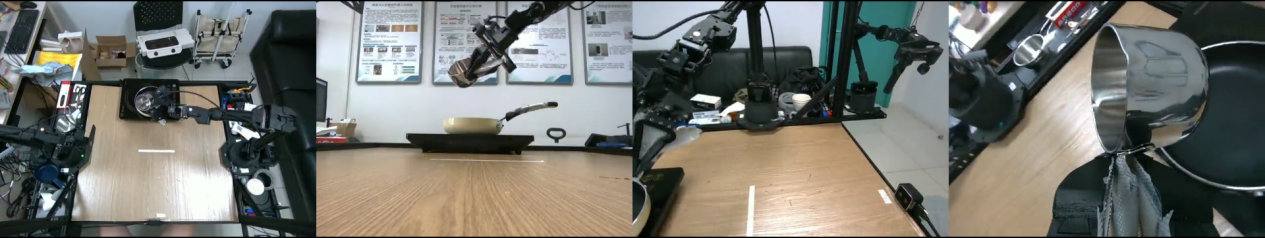}
        \caption{Continuous pouring issue 2}
        \label{fig:pouring_failure_2}
    \end{figure}
    \begin{figure}[H]
        \centering
        \includegraphics[width=0.8\linewidth]{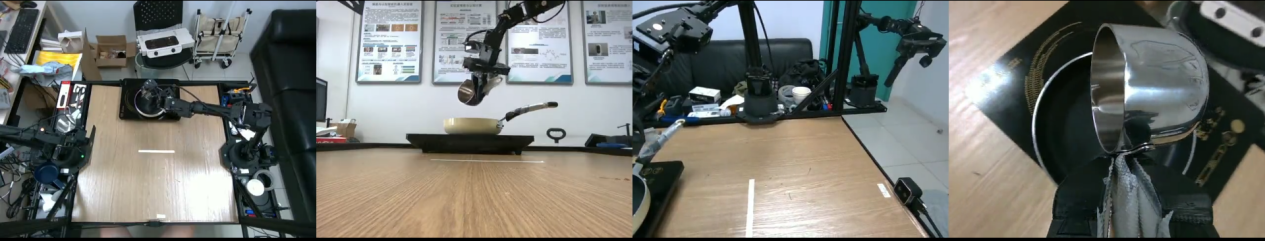}
        \caption{Continuous pouring issue 3}
        \label{fig:pouring_failure_3}
    \end{figure}
    Due to the time symmetry of the task, the act algorithm lacking timing information cannot distinguish whether it is in the stage of preparing to pour ingredients or preparing to put the small pot back, thus leading to continuous pouring without returning the pot after pouring ingredients.

     \item Failure(Unstable grasping): 
    \begin{figure}[H]
        \centering
        \includegraphics[width=0.8\linewidth]{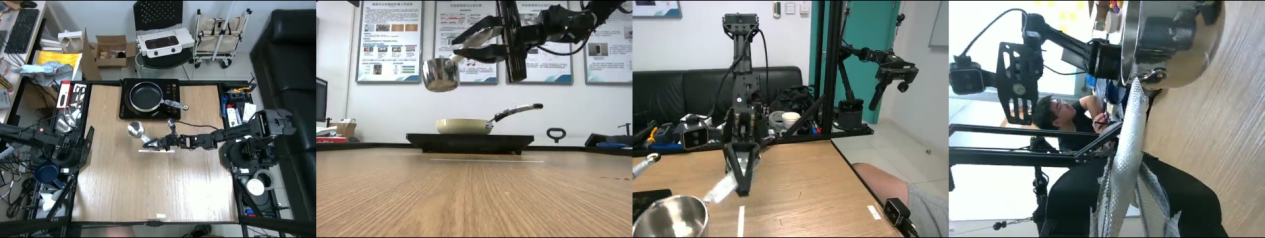}
        \caption{Unstable grasping 1}
        \label{fig:grasping_failure_1}
    \end{figure}
    \begin{figure}[H]
        \centering
        \includegraphics[width=0.8\linewidth]{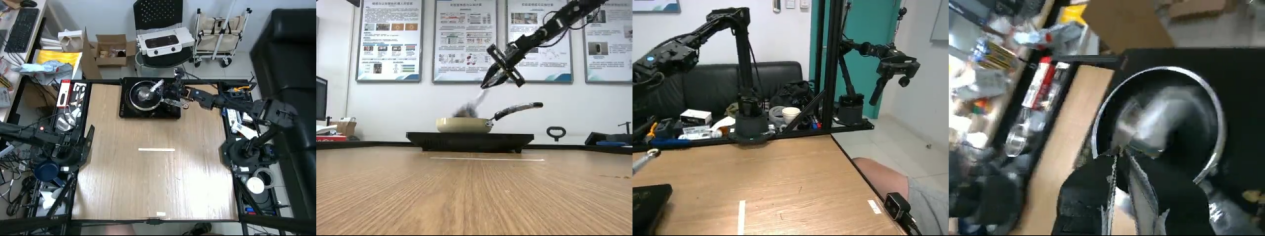}
        \caption{Unstable grasping 2}
        \label{fig:grasping_failure_2}
    \end{figure}
    The handle of the small pot is thin, so when the robotic arm grasps the pot, the pot is prone to tilt, causing food to fall out or even the entire small pot to fall off.
\end{itemize}

\begin{figure}[H]
    \centering
    \includegraphics[width=0.5\textwidth]{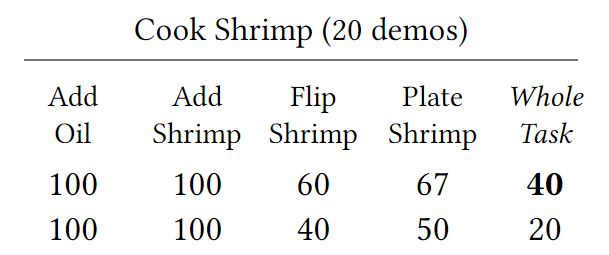}
    \caption{Success Rates for Each Step of the Original Paper's Shrimp Stir-frying Task}
    \label{fig:paper_success2}
\end{figure}

\begin{figure}[H]
    \centering
    \includegraphics[width=0.5\textwidth]{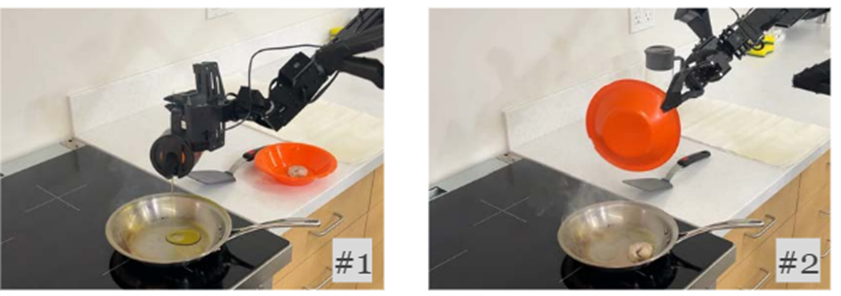}
    \caption{Original Paper's Photo of Pouring Ingredients into Pot}
    \label{fig:paper_scene2}
\end{figure}

\textbf{5. Comparison with Original Paper} \\
The original paper's success rates for pouring oil into the pot and adding shrimp were both 100\% (see Figure \ref{fig:paper_success2}), higher than our experimental success rate. However, since the objects being poured into the pot and the pot itself are different, we consider this gap reasonable (see Figure \ref{fig:paper_scene2}).


\textbf{6. Summary} \\
i. There is a problem in this link: the small pot often cannot be retrieved after pouring, and if ingredients are put in, the small pot may fall into the pan and contaminate the ingredients.

ii. There will be a problem of pouring twice in a row.

iii. The execution speed and completeness of the task need to be improved. In subsequent data collection, the symmetry of the action should be eliminated, for example, placing the small pot in another place after pouring.

\subsection*{III. Stir-frying with Spatula}
\textbf{1. Action Flow} \\
The right robotic arm picks up the spatula placed in the middle of the table, stirs the food in the pot with the spatula, and finally takes the spatula out.

\begin{figure}[H]
    \centering
    \includegraphics[width=0.8\linewidth]{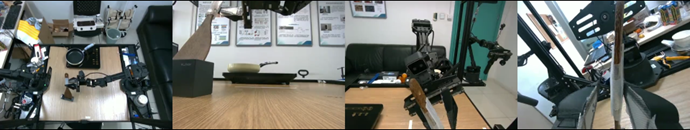}
    \caption{Stir-frying step 1}
    \label{fig:stir_fry_1}
\end{figure}
\begin{figure}[H]
    \centering
    \includegraphics[width=0.8\linewidth]{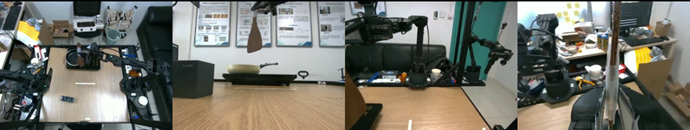}
    \caption{Stir-frying step 2}
    \label{fig:stir_fry_2}
\end{figure}
\begin{figure}[H]
    \centering
    \includegraphics[width=0.8\linewidth]{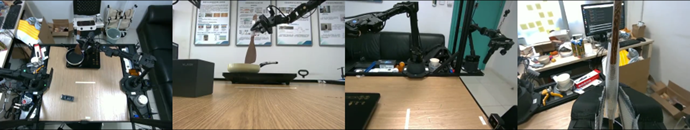}
    \caption{Stir-frying step 3}
    \label{fig:stir_fry_3}
\end{figure}
\begin{figure}[H]
    \centering
    \includegraphics[width=0.8\linewidth]{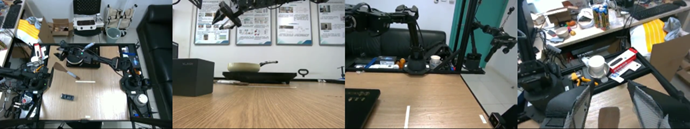}
    \caption{Stir-frying step 4}
    \label{fig:stir_fry_4}
\end{figure}

\textbf{2. Data Collection} \\
A total of 50 data samples were collected, each with a duration of 15 seconds. Since the spatula is relatively flat, a shelf was used to stand the spatula up for easy grasping. The frying pan was placed on the induction cooker. When stir-frying, the spatula was kept as close to the bottom of the pan as possible to ensure sufficient stirring. After taking out the spatula, it was placed on the table because putting it back on the shelf required overly precise movements.

\textbf{3. Model Training} \\
50 data samples were used for 8000 training iterations.

\textbf{4. Execution and Inference} \\
The best-performing model was selected for final inference. 10 experiments were conducted. The success rate of grasping the spatula and putting it into the pot was 90\%, but stir-frying actions rarely occurred, and the success rate of taking the spatula out of the pot was 60\%.

\begin{itemize}
    \item Success: 
    \begin{figure}[H]
        \centering
        \includegraphics[width=0.8\linewidth]{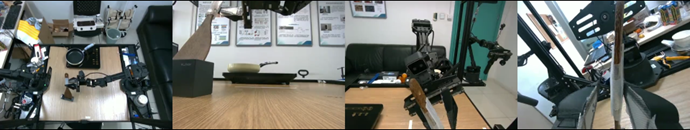}
        \caption{Successful spatula grasping}
        \label{fig:spatula_success_1}
    \end{figure}
    \begin{figure}[H]
        \centering
        \includegraphics[width=0.8\linewidth]{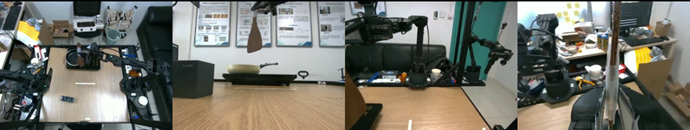}
        \caption{Successful spatula placement in pan}
        \label{fig:spatula_success_2}
    \end{figure}
    \begin{figure}[H]
        \centering
        \includegraphics[width=0.8\linewidth]{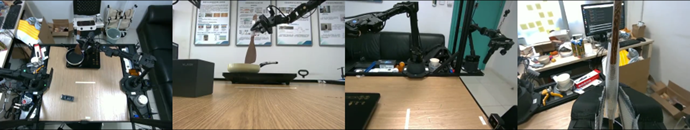}
        \caption{Successful stirring}
        \label{fig:spatula_success_3}
    \end{figure}
    \begin{figure}[H]
        \centering
        \includegraphics[width=0.8\linewidth]{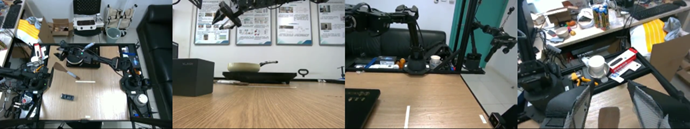}
        \caption{Successful completion}
        \label{fig:spatula_success_4}
    \end{figure}
    
    \item Failure(Inability to grasp): 
    \begin{figure}[H]
        \centering
        \includegraphics[width=0.8\linewidth]{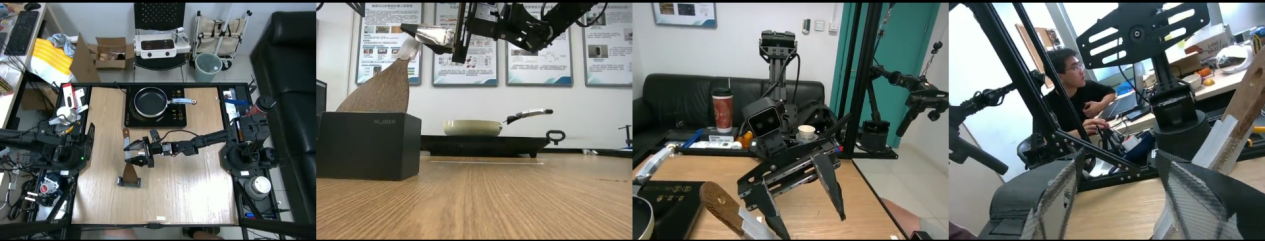}
        \caption{Failed grasping attempt}
        \label{fig:spatula_failure_1}
    \end{figure}
    \begin{figure}[H]
        \centering
        \includegraphics[width=0.8\linewidth]{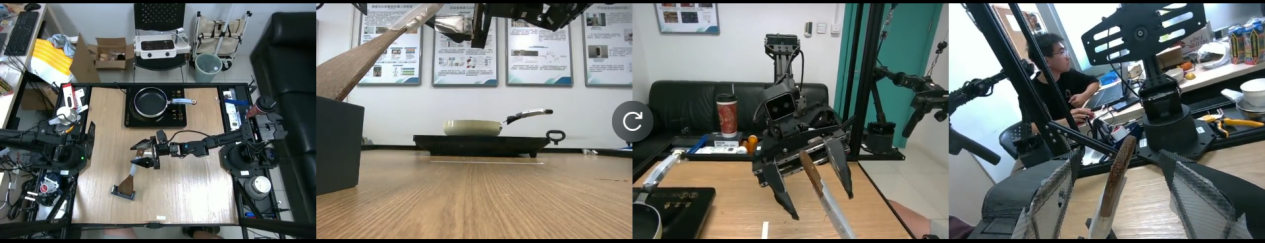}
        \caption{Incomplete stirring}
        \label{fig:spatula_failure_2}
    \end{figure}
    In addition, the main difficulty of this task is that due to the limitation of the act algorithm, it is difficult to train large-scale food stirring movements. In almost every inference, the robotic arm could not use the spatula to stir-fry in the pot well.
\end{itemize}

\textbf{5. Comparison with Original Paper} \\
In the original paper, without joint training with static ALOHA collected data, the success rate of flipping shrimp was 40\% (see Figure \ref{fig:paper_success2}), while our success rate of flipping food was also low at 10\%. The better performance of the original paper may be related to the fact that the shrimp they used were easier to flip. In addition, our collected data involved back-and-forth stirring of food, which was more difficult to train than the single flipping action in the original paper.

\textbf{6. Summary} \\
i. The stir-frying action is the biggest problem in this link. The analysis shows that this is because the displacement of this action is not obvious, leading to inadequate learning.

ii. The action of taking the spatula out of the pot and placing it on the table is not elegant and needs improvement in the future.

iii. In the future, the stir-frying action can be changed from back-and-forth translation to stirring in the pot. Preliminary analysis shows that this will not be smoothed out by the act algorithm.

\subsection*{IV. Pouring Dish from Pot to Plate}
\textbf{1. Action Flow} \\
The right robotic arm grasps the handle of the pot, pours the food in the pot into the plate located in the middle of the table, and then places the pot back on the induction cooker.

\begin{figure}[H]
    \centering
    \includegraphics[width=0.8\linewidth]{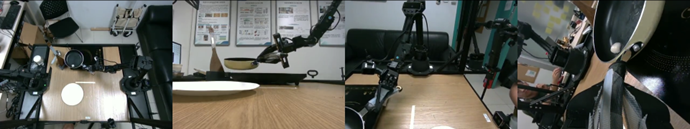}
    \caption{Pouring to plate step 1}
    \label{fig:plate_pour_1}
\end{figure}
\begin{figure}[H]
    \centering
    \includegraphics[width=0.8\linewidth]{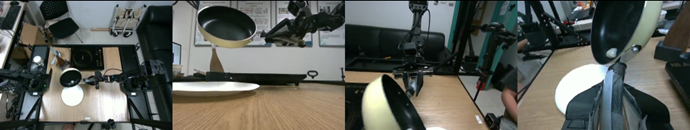}
    \caption{Pouring to plate step 2}
    \label{fig:plate_pour_2}
\end{figure}
\begin{figure}[H]
    \centering
    \includegraphics[width=0.8\linewidth]{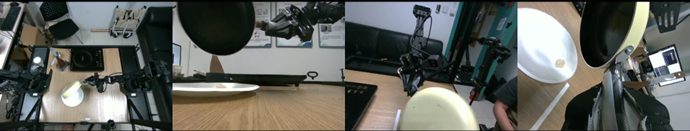}
    \caption{Pouring to plate step 3}
    \label{fig:plate_pour_3}
\end{figure}
\begin{figure}[H]
    \centering
    \includegraphics[width=0.8\linewidth]{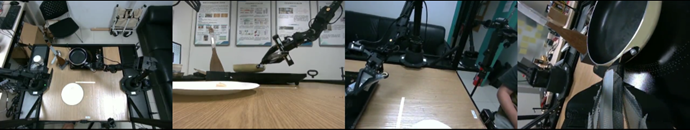}
    \caption{Pouring to plate step 4}
    \label{fig:plate_pour_4}
\end{figure}

\textbf{2. Data Collection} \\
A total of 100 data samples were collected in two batches, and the latter 50 samples were used after comparison, each with a duration of 20 seconds. When grasping the pot handle, it was better to grasp the part close to the pot body so that the tail was stuck on the robotic arm for better stability. When pouring the dish, the pot was flipped as completely as possible to pour out all the food.

\textbf{3. Model Training} \\
The latter 50 data samples were used for 8000 training iterations.

\textbf{4. Execution and Inference} \\
The best-performing model was selected for final inference. 10 experiments were conducted, and the success rate of completing the action completely reached 80\%. There were few cases of successful grasping but unstable holding. The pot was accurately located and grasped by the gripper every time, and it was held firmly in most cases. Sometimes, the robotic arm could not be retracted, which was the same as in the first stage.

\begin{itemize}
    \item Success: 
    \begin{figure}[H]
        \centering
        \includegraphics[width=0.8\linewidth]{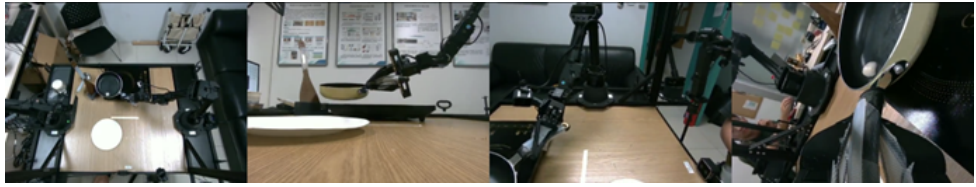}
        \caption{Successful pouring to plate step 1}
        \label{fig:plate_success_1}
    \end{figure}
    \begin{figure}[H]
        \centering
        \includegraphics[width=0.8\linewidth]{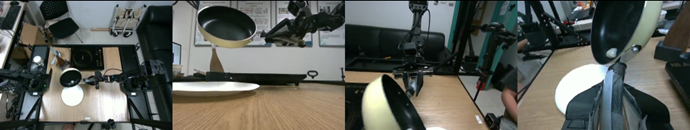}
        \caption{Successful pouring to plate step 2}
        \label{fig:plate_success_2}
    \end{figure}
    \begin{figure}[H]
        \centering
        \includegraphics[width=0.8\linewidth]{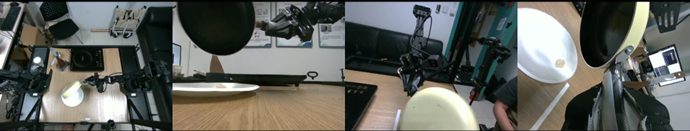}
        \caption{Successful pouring to plate step 3}
        \label{fig:plate_success_3}
    \end{figure}
    \begin{figure}[H]
        \centering
        \includegraphics[width=0.8\linewidth]{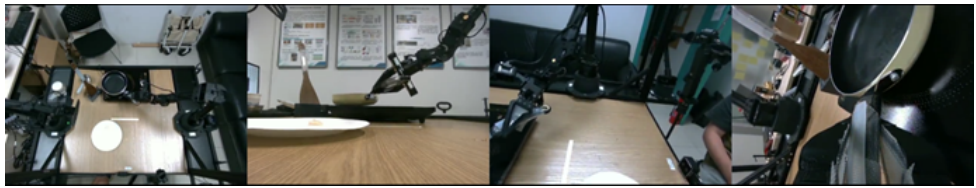}
        \caption{Successful completion}
        \label{fig:plate_success_4}
    \end{figure}
    
    \item Failure(Inability to find the frying pan): 
    \begin{figure}[H]
        \centering
        \includegraphics[width=0.8\linewidth]{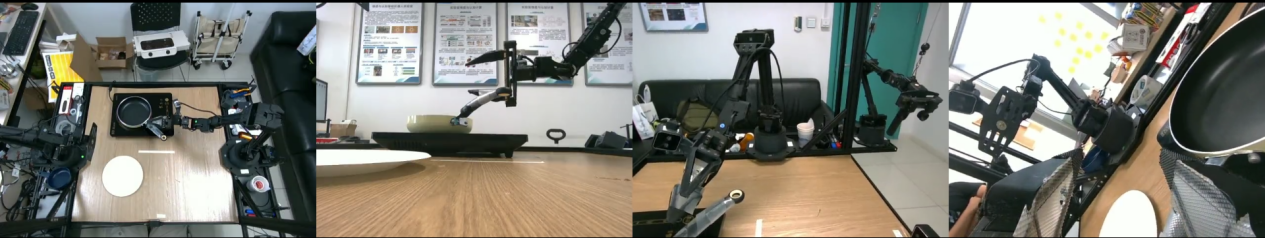}
        \caption{Failed to locate pan}
        \label{fig:plate_failure_1}
    \end{figure}
    The data collection was too rapid, resulting in failure to learn the positioning process well.
\end{itemize}

\begin{figure}[htbp]
    \centering
    \includegraphics[width=0.5\textwidth]{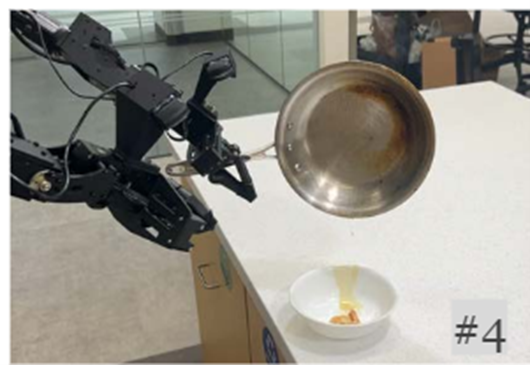}
    \caption{Original Paper's Photo of Pouring Out Ingredients}
    \label{fig:paper_scene3}
\end{figure}

\textbf{5. Comparison with Original Paper} \\
In the original paper, without joint training with static ALOHA collected data, the success rate for removing food from the pot was 50\% (see Figure \ref{fig:paper_success2}), lower than our success rate. The reason may be that the original paper used a small bowl to hold food (see Figure \ref{fig:paper_scene3}), while we used a plate, which may be less difficult than the original paper, thus resulting in a higher success rate.

\textbf{6. Summary} \\
i. The completion of this stage was poor after the first sampling. The analysis showed that the sampling was too fast, preventing ALOHA2 from learning the positioning process, thus leading to inability to grasp.

ii. However, the overall action completion is slow at this stage, which is related to the slower speed during the second sampling to achieve more accurate positioning.

iii. The speed of this link needs to be improved in the future without reducing stability.

iv. In addition, similar to the second stage, secondary pouring may occur in this stage, so changing the final position of the pot can be considered to avoid this problem.

\subsection*{V. Comparison of Overall Results with Original Paper}
\begin{figure}[H]
    \centering
    \includegraphics[width=0.8\linewidth]{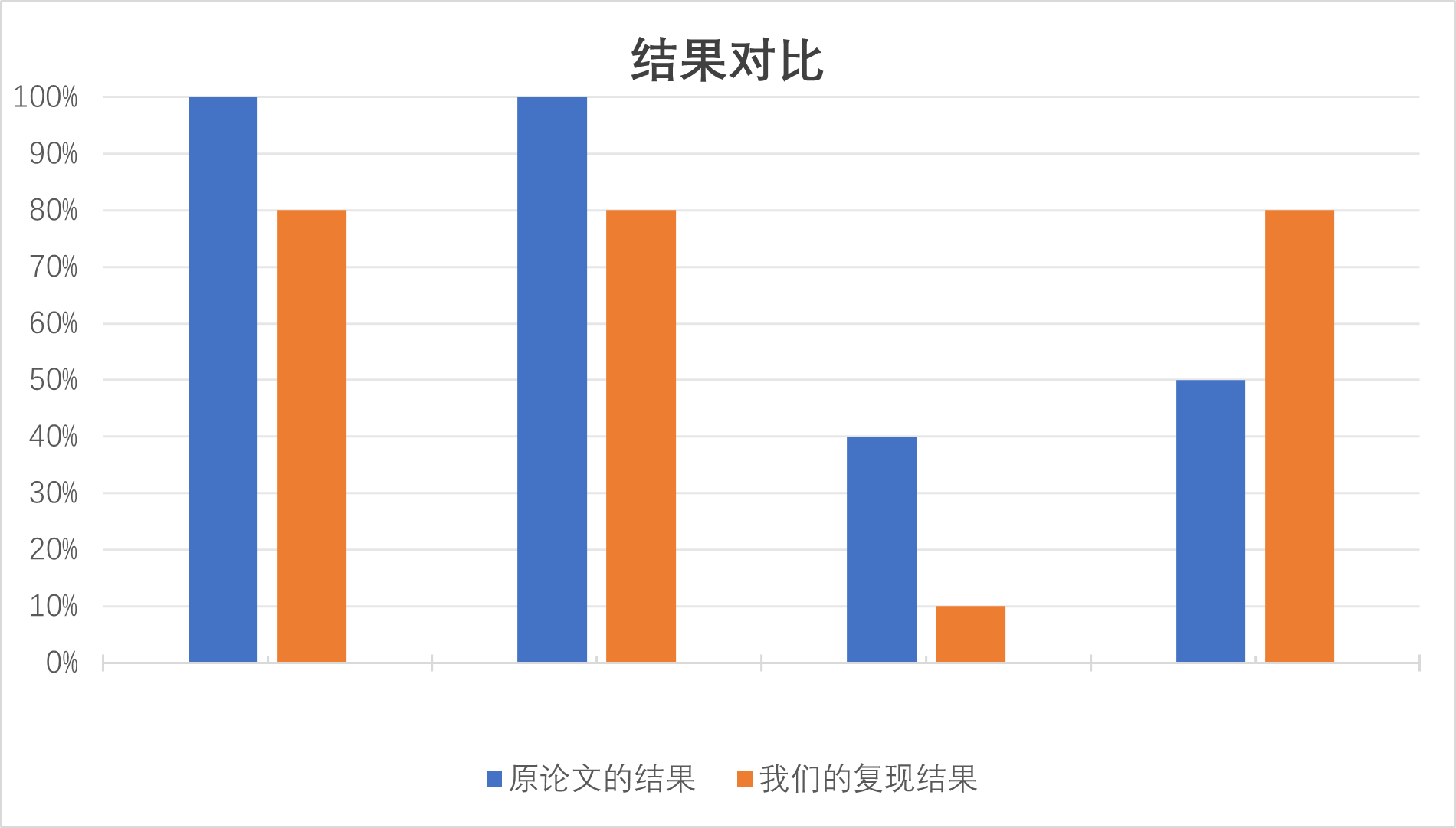}
    \caption{Overall performance comparison with original paper}
    \label{fig:overall_comparison}
\end{figure}
It can be seen that our reproduction results are quite close to the original paper, although the results are slightly different due to some specific experimental settings and data collection differences.

\bibliographystyle{ieeetr}
\bibliography{references}

\begin{thebibliography}{1}

\bibitem{zhao2024aloha2}
T.~Z. Zhao, Z.~Fu, C.~Finn, P.~Abbeel, {\em et~al.}, ``Aloha 2: An enhanced low-cost hardware for bimanual teleoperation,'' {\em arXiv preprint arXiv:2405.02292}, 2024.

\bibitem{zhao2023aloha}
T.~Z. Zhao, Z.~Fu, A.~Murali, K.~Lee, and P.~Abbeel, ``Learning fine-grained bimanual manipulation with low-cost hardware,'' {\em arXiv preprint arXiv:2304.13705}, 2023.

\bibitem{brohan2022rt1}
A.~Brohan, N.~Brown, Y.~Chen, and et~al., ``Rt-1: Robotics transformer for real-world control at scale,'' {\em arXiv preprint arXiv:2212.06817}, 2022.

\bibitem{brohan2023rt2}
A.~Brohan, N.~Brown, Y.~Chen, and et~al., ``Rt-2: Vision-language-action models transfer web knowledge to robotic control,'' {\em arXiv preprint arXiv:2307.15818}, 2023.

\bibitem{openxembodiment2024dataset}
O.~X.-E. Team, ``Open x-embodiment: Robotic learning datasets and rt-x models,'' in {\em Conference on Robot Learning (CoRL)}, 2024.

\bibitem{dann2021robot}
C.~Dann and V.~Kr{\"u}ger, ``Robot cooking with learning from demonstration: A review,'' {\em Robotics and Autonomous Systems}, vol.~144, p.~103841, 2021.

\end{thebibliography}

\end{document}